\documentclass[letterpaper]{article} 
\usepackage{aaai2026}
\usepackage{times}  
\usepackage{helvet}  
\usepackage{courier}  
\usepackage[hyphens]{url}  
\usepackage{graphicx} 
\usepackage{natbib}  
\usepackage{caption} 
\usepackage{algorithm}
\usepackage{algorithmic}

\usepackage{newfloat}
\usepackage{listings}
\DeclareCaptionStyle{ruled}{labelfont=normalfont,labelsep=colon,strut=off} 
\floatstyle{ruled}
\newfloat{listing}{tb}{lst}{}
\floatname{listing}{Listing}
\usepackage{amsfonts}
\usepackage{amssymb}
\usepackage{amsmath}
\usepackage{amsthm}
\usepackage{graphicx}
\usepackage{xcolor}

\definecolor{firebrick}{RGB}{178,34,34}

\title{Revisiting Data Attribution for Influence Functions}
\author {
    Hongbo Zhu, Angelo Cangelosi
}
\affiliations {
    Manchester Centre for Robotics and AI, The University of Manchester\\
    \texttt{\{hongbo.zhu, angelo.cangelosi\}@manchester.ac.uk}
}

\usepackage{bibentry}

\nocopyright
\begin{document}

\maketitle

\begin{abstract}
The goal of data attribution is to trace the model's predictions through the learning algorithm and back to its training data. thereby identifying the most influential training samples and understanding how the model's behavior leads to particular predictions. Understanding how individual training examples influence a model’s predictions is fundamental for machine learning interpretability, data debugging, and model accountability. Influence functions, originating from robust statistics, offer an efficient, first-order approximation to estimate the impact of marginally upweighting or removing a data point on a model’s learned parameters and its subsequent predictions, without the need for expensive retraining. This paper comprehensively reviews the data attribution capability of influence functions in deep learning. We discuss their theoretical foundations, recent algorithmic advances for efficient inverse-Hessian-vector product estimation, and evaluate their effectiveness for data attribution and mislabel detection. Finally, highlighting current challenges and promising directions for unleashing the huge potential of influence functions in large-scale, real-world deep learning scenarios.
\end{abstract}

\section{Introduction}

The rapid growth of datasets required to train state-of-the-art deep learning models has outpaced the ability of humans to carefully curate them, leading to significant issues such as outliers and label misannotations \cite{yeh2018representer}. These problems commonly arise in settings where training data is scraped from the web and annotated using heuristics or non-expert crowd workers, often under budget constraints that preclude quality assurance \cite{ghorbani2019data}. As a result, understanding and improving model behavior increasingly necessitates tracing predictions back to the specific training examples that shaped them. A principled way to formalize this is to consider the counterfactual impact of individual training points: how would a model's prediction change if a particular example were altered or removed? Directly answering this by leave-one-out retraining \cite{hammoudeh2024training} is typically computationally infeasible, which motivates the use of Influence Functions(IFs), a classical method from robust statistics \cite{hampel1974influence} that estimates how an infinitesimal upweighting of a training point affects model parameters and, consequently, predictions. IFs provide a practical and theoretically grounded approach to data attribution, enabling researchers to identify which training examples most influenced a model’s output on a given test input, and offering a lens through which to interpret and debug deep neural networks.

\section{Related Works}
The resurgence of interest in influence functions (IFs) within deep learning began with the work of \citet{koh2017understanding}, who introduced practical algorithms for estimating the effect of individual training points on test predictions using first- and second-order gradient information. This eliminated the need for retraining models from scratch and enabled data attribution for moderately sized models and datasets. As a result, IFs quickly became valuable tools for model debugging \cite{liu2022debugging}, data curation \cite{agia2025cupid}, bias detection \cite{wang2024fairif}, and the identification of mislabeled or adversarial data points \cite{cohen2020detecting}.

However, the widespread adoption of IFs has been hindered by their computational cost, particularly the need to calculate and invert high-dimensional Hessian matrices. To mitigate these limitations, several improvements have been proposed. FASTIF \cite{guo2020fastif} accelerates IF computation through a combination of KNN-based preselection and parallelization, achieving substantial speedups with minimal loss in attribution fidelity. TRAK \cite{park2023trak} leverages insights from neural tangent kernels to approximate influence scores using a small number of model checkpoints, enabling scalable and accurate attribution across vision, language, and multimodal models. Similarly, DataInf \cite{kwon2023datainf} provides a closed-form solution for parameter-efficient finetuning scenarios such as LoRA \cite{hu2022lora}, making it suitable for large language models (LLMs) and diffusion models in practice.

Despite these advancements, the applicability of IFs in deep, non-convex settings remains an open challenge.  \citet{basu2020influence} and \citet{bae2022if} have highlighted discrepancies between IF estimates and exact leave-one-out retraining in neural networks, arising from non-convexity, initialization variance, and optimization dynamics. Nonetheless, influence scores may still correlate with alternative measures like the Proximal Bregman Response Function (PBRF), preserving interpretability in many cases. To improve robustness, \citet{bae2024training} introduced hybrid approaches, SOURCE that combine implicit differentiation with unrolling, offering greater accuracy under underspecification, optimizer bias, and multi-stage training regimes.

While influence functions and their extensions have become essential tools for model interpretability, key challenges remain in scaling, robustness under non-convex optimization, and adapting to new model classes and training paradigms. Current research continues to address these through approximation, hybridization with unrolling, and kernel-based innovations, pushing the boundary for practical data attribution in modern deep learning.

\section{Preliminaries}

In supervised learning, let the input space be  \(\mathcal{X} \subseteq \mathbb{R}^{d} \) and the output space be \(\mathcal{Y}\), a training set $Z = \{z_1, \ldots, z_n\}$, where each data point \(z_{i}=\left(x_{i}, y_{i}\right) \in\) \(\mathcal{X} \times \mathcal{Y}\). For a given data point \(z\) and model parameters \(\theta \in \mathbb{R}^{p}\), let \(L(z, \theta)\) denote the loss function. The empirical risk is defined as \(\frac{1}{n} \sum_{i=1}^{n} L\left(z_{i}, \theta\right)\)  and the empirical risk minimizer is given by \(\hat{\theta} \stackrel{\text { def }}{=} \arg \min _{\theta \in \Theta} \frac{1}{n} \sum_{i=1}^{n} L\left(z_{i}, \theta\right)\), we assume that the empirical risk is twice differentiable and strictly convex in \(\hat{\theta}\), ensuring that $\hat{\theta}$ is unique so it's Hessian is positive definite.

\subsection{Data Attribution}
Data attribution aims to quantify how much each training example contributes to a specific prediction made by a learned model. Given a trained model $f(z; \hat{\theta})$, a data attribution method assigns an importance score to each training point in relation to its influence on a particular test example \cite{park2023trak}. Formally, a data attribution method is defined as a function:
\[
\tau(z, Z):z \times \mathcal{Z} \to \mathbb{R}^n 
\]
which, given a test input $z$ and a training set $Z$ , returns a vector of real-valued scores. Each entry in this vector represents the contribution of a training example $z_i \in Z$ to the model's output $f(z; \hat{\theta})$. A straightforward yet impractical approach to data attribution is to assess how the model's parameters change when a single training point $z$ is removed. This yields a new estimator:
\[
\hat{\theta}_{-z} \stackrel{\text { def }}{=} \arg \min _{\theta \in \Theta} \frac{1}{n}\sum_{z_{i} \neq z} L\left(z_{i}, \theta\right)
\]
The parameter difference \(\hat{\theta}_{-z}-\hat{\theta}\) then reflects the influence of the removed training point.
However, this brute-force method requires retraining the model $n$ times, one for each training point, which is computationally prohibitive for large-scale datasets. More efficient data attribution methods have therefore been developed to approximate these influence scores without full retraining.

\subsection{Influence Function}

To address this inefficiency, the influence function offers a first-order approximation to the effect of removing a training point on the learned parameters. Let $\hat{\theta}_{\epsilon, z}$ be the model parameters obtained by upweighting a training example $z$ by an infinitesimal amount $\epsilon$, \(\hat{\theta}_{\epsilon, z} \stackrel{\text { def }}{=} \arg \min _{\theta \in \Theta} \frac{1}{n} \sum_{i=1}^{n} L\left(z_{i}, \theta\right)+\epsilon L(z, \theta)\). A classic result from \citet{cook1982residuals} tells us that the influence of up-weighting \(z\) on the parameters \(\hat{\theta}\) is given by

\begin{align}\label{eq:1}
\left.\mathcal{I}_{\text {params }}(z) \stackrel{\text { def }}{=} \frac{d \hat{\theta}_{\epsilon, z}}{d \epsilon}\right|_{\epsilon=0}=-H_{\hat{\theta}}^{-1} \nabla_{\theta} L(z, \hat{\theta})
\end{align}

Where \(H_{\hat{\theta}} \stackrel{\text { def }}{=} \frac{1}{n} \sum_{i=1}^{n} \nabla_{\theta}^{2} L\left(z_{i}, \hat{\theta}\right)\) is the Hessian of the empirical risk at $\hat{\theta}$, assumed to be positive definite (PD) to ensure invertibility (all eigenvalues $\lambda$ are positive). Which essentially requires $H_{\hat{\theta}}$ to be unique optimal, if $\hat{\theta}$ is not at the optimal, the Hessian may have negative values. see Appendix 1.1 for the derivation of $\mathcal{I}_{\text {params }}(z)$. Since removing a point \(z\) is approximately equivalent to upweighting it by \(\epsilon=-\frac{1}{n}\), we can linearly approximate the parameter change caused by removing \(z\) without retraining the model by computing \(\hat{\theta}_{-z}-\hat{\theta} \approx-\frac{1}{n} \mathcal{I}_{\text {params }}(z)\), the proof can be seen in Appendix 1.2. 

Next, we can propagate the effect of this parameter change to a test loss using the chain rule. Specifically, the influence of upweighting \(z\) on the loss at a test point \(z_{\text {test }}\) is:

\begin{equation} 
\begin{aligned}\label{eq-2}
\mathcal{I}_{\text{loss}}\left(z, z_{\text{test}}\right)
& \stackrel{\text{def}}{=}
\left.\frac{d L\left(z_{\text{test}}, \hat{\theta}_{\epsilon, z}\right)}{d \epsilon}\right|_{\epsilon=0} \\
& = -\nabla_{\theta} L\left(z_{\text{test}}, \hat{\theta}\right)^{\top}
H_{\hat{\theta}}^{-1} \nabla_{\theta} L(z, \hat{\theta}) \in \mathbb{R}
\end{aligned}
\end{equation}

$\mathcal{I}_{\text{loss}}(z, z_{\text{test}})$ measures the change in loss at the test point $z_{\text{test}}$ , caused by up-weighting the training point $z$  by an infinitesimally small amount. Here, $\nabla_{\theta} L\left(z_{\text{test}}, \hat{\theta}\right)^{\top} \in \mathbb{R}^{1 \times p} $, $H_{\hat{\theta}}^{-1} \in \mathbb{R}^{p \times p}$ and $\nabla_{\theta}L(z, \hat{\theta}) \in \mathbb{R}^p$, resulting in a scalar quantity after multiplication. The sign of $\mathcal{I}_{\text{loss}}(z, z_{\text{test}})$ is informative:

\begin{itemize}
    \item If it is \textbf{positive}, upweighting the training point $z$ increases the test loss, implying that $z$ is harmful to the performance on $z_{\text{test}}$.
    \item If it is \textbf{negative}, upweighting that of $z$ reduces the test loss, implying that $z$ is helpful to the prediction at $z_{\text{test}}$.
\end{itemize}


\subsection{Gauss-Newton Hessian}
In theory, influence functions are derived based on the assumption that the model has been trained to a global minimum denoted by $\hat{\theta}$. But in practice, we usually get the parameter $\theta$ by running SGD \cite{amari1993backpropagation} with early stopping or on non-convex objectives (not fully trained to convergence), so $\theta \not = \hat{\theta}$, resulting the true $H_{\theta}$ ofen singular or indefinite (has negative eigenvalues). This violates the positive definiteness assumption needed for influence function derivations that rely on the inverse Hessian. Thus motivates the use of approximations like the Gauss-Newton Hessian (GNH) \cite{botev2017practical} namely $G$ \footnote{GNH can be seen as an approximation to $H$ which linearizes the network’s parameter-output mapping around the current parameters, and the dimension of $G$ is the same as $H$ of $p \times p$.} and damped versions $(G + \lambda I)$ that are guaranteed to be positive definite, even when the model isn’t fully trained or the loss landscape is non-convex. The GNH is always positive semidefinite, when a damping term $\lambda I$ is added to $G$, $(G + \lambda I)$ becomes positive definite, ensuring stability and invertibility in iHVPs (inverse-Hessian-vector products) \cite{dagréou2024howtocompute}. This makes it a more robust substitute for $H$ under practical deep learning settings, applying the above changes yields the following update:

\begin{align}\label{eq-3}
\mathcal{I_{\text{loss}}}(z,z_{\text{test}}) =-\nabla_\theta L(z_{\text{test}},\theta)^\top{(G + \lambda I)}^{-1} \nabla_{\theta} L(z, \theta)
\end{align}

\section{Computation Challenge}

Computing the influence score defined in Eq.~\eqref{eq-3} for a given test point \( z_{\text{test}} \) involves two primary steps. First, one must compute the inverse Hessian-vector product (IHVP):
\(
v = (G + \lambda I)^{-1} \nabla_\theta L(z_{\text{test}}, \theta),
\)
where \( G \) denotes the Gauss-Newton approximation to the Hessian and \( \lambda I \) is a damping term that ensures positive definiteness. Second, for each training point \( z_i \in Z \), the influence score is computed via a dot product:
\(
\mathcal{I}_{\text{loss}}(z_i, z_{\text{test}}) = -\nabla_\theta L(z_i, \theta)^\top v.
\)
This procedure entails \( O(n) \) computations per test example, which becomes computationally intensive for large-scale datasets and high-dimensional parameter spaces.

Regardless of whether it is necessary to calculate the influence value for the entire training dataset, the first challenge alone is very tricky. As the computation of \( (G + \lambda I)^{-1} \) is computationally prohibitive in modern deep networks. Specifically, for a model with \( n \) training examples and \( p \)-dimensional parameters, forming the matrix \( G \in \mathbb{R}^{p \times p} \) requires \( \mathcal{O}(np^2) \) operations, and matrix inversion incurs a cost of \( \mathcal{O}(p^3) \). Additionally, storing the full matrix requires \( \mathcal{O}(p^2) \) memory, which is infeasible for models with millions, even billions of parameters. To address this limitation, recent methods employ \emph{implicit Hessian-vector products (HVPs)}~\cite{dagréou2024howtocompute}, which allow efficient computation and expressions of the form \( Gv \) without explicitly forming \( G \). Based on this idea, two widely-used techniques for approximating IHVPs are:

\begin{itemize}
    \item \textbf{LISSA}~\cite{agarwal2017second}: A stochastic algorithm that approximates the inverse using a truncated Neumann series expansion, allowing linear-time computation relative to the number of iterations.
    \item \textbf{EK-FAC}~\cite{grosse2023studying}: An eigenvalue-corrected Kronecker-factored curvature approximation that efficiently estimates inverse curvature by factorizing and adjusting the diagonal structure of the Fisher information matrix.
\end{itemize}

These methods enable scalable and memory-efficient estimation of influence scores without requiring explicit inversion of large Hessians.

\section{Efficient IHVP Approximation}

\subsection{LiSSA} \label{LiSSA}

LiSSA approximates the IHVP via a lightweight stochastic recursion.  Starting from the initial guess $\mathbf{r}_0 = \mathbf{v}$, the update rule \(\mathbf{r}_{j}= \mathbf{v}+ \bigl(I - \alpha(\tilde{G} + \lambda I)\bigr)\mathbf{r}_{j-1}\)is applied at every iteration, where $\tilde{G}$ is an unbiased stochastic mini-batch estimate of the Hessian $G$, the term $\lambda I$ provides Tikhonov regularisation, and the step size $\alpha>0$ is chosen so that $\alpha(\tilde{G}+\lambda I)\preceq I$.  Under this spectral condition, $\mathbf{r}_J$ converges to $\alpha^{-1}(G+\lambda I)^{-1}\mathbf{v}$  after $J$ iterations, furnishing an accurate surrogate of the desired inverse product. Each iteration requires only one Hessian–vector product, incurring a cost of $\mathcal{O}(bp)$ operations for for mini-batch of size of $b$; overall, the total cost is $\mathcal{O}(J\cdot bp)$ operations and $\mathcal{O}(p)$ memory. Compared to traditional methods like conjugate gradient \citep{martens2010deep}, which assume full-batch positive-definite matrices. By exploiting mini-batch curvature estimates, LiSSA enables practical second-order sensitivity analysis in contemporary deep networks while maintaining linear-time efficiency. However, LiSSA is still a computationally expensive algorithm because each Hessian-vector product (HVP) is at least as costly as computing a gradient, and it often requires thousands of iterations to produce accurate results.

\subsection{EK-FAC} \label{EK-FAC}
To efficiently approximate inverse Hessian-vector products (IHVPs) for influence function estimation, we adopt structured approximations to the $G$. K-FAC exploits the layer-wise structure of neural networks to approximate the $G$. For an MLP with input activations $a_{l-1} \in \mathbb{R}^{d_{l}}$, weights $W_{l} \in \mathbb{R}^{p_{l}\times d_{l}}$ and activation function $\phi$ , the pre-activation output is $y_{l}=W_{l}a_{l-1}$ then $a_{l}=\phi(y_{l})$. The parameter pseudo-gradient for layer \( l \)  given by $D{W_{l}} = D_{y_l}a_{l-1}^\top$ and can be further expressed as a Kronecker product: \(D{\theta_l}= {a}_{l-1} \otimes D_{y_l}\), where $\theta_l = \text{vec}(W_l)$, which converting a 2D matrix into a 1D vector by flattening it column-by-column and $\otimes$ denotes the Kronecker product. This leads to a block-diagonal GNH approximation per layer:

\begin{equation}
\begin{aligned}
G_l 
&= \mathbb{E}[\mathcal{D} {\theta}_l \mathcal{D} {\theta}_l^\top] 
= \mathbb{E}[{a}_{l-1} {a}_{l-1}^\top \otimes \mathcal{D} y_l \mathcal{D} y_l^\top] \\
&\approx \mathbb{E}[{a}_{l-1} {a}_{l-1}^\top] \otimes \mathbb{E}[\mathcal{D} y_l \mathcal{D} y_l^\top] = A_{l-1} \otimes Y_l.
\end{aligned}
\end{equation}

Where \( A_{l-1} \) and \( Y_l \) are uncentered covariances of inputs and pre-activation pseudo-gradients, respectively. Using Kronecker product identities, the IHVP for vector \( v_l \) is computed efficiently:

\begin{equation}
\begin{aligned}
G_l^{-1} v_l 
\approx (A_{l-1} \otimes Y_l)^{-1} v_l =(A_{l-1}^{-1} \otimes Y_l^{-1}) v_l \\
= \mathrm{vec}(Y_l^{-1} V_l A_{l-1}^{-1})
\end{aligned}
\end{equation}

where \( V_l \in \mathbb{R}^{p_{l} \times d_{l}} \) is the reshaped version of \( v_l \). This reduces the inversion cost to that of two smaller matrices. K-FAC assumes the eigenvalues of \( G = A \otimes Y \) factor as \( \Lambda_A \otimes \Lambda_Y \). 

EK-FAC relaxes this assumption by directly estimating variances in the Kronecker eigenspace. Given eigendecompositions:
\(A = Q_A \Lambda_A Q_A^\top,\quad Y = Q_Y \Lambda_Y Q_Y^\top\),
EK-FAC rewrites the approximation as:
\(G \approx (Q_A \otimes Q_Y) \Lambda (Q_A \otimes Q_Y)^\top\), 
where $Q_A, Q_S$ are orthonormal eigenvector matrices of Kronecker factors $A$ and $S$, $\Lambda$ is a diagonal matrix of fitted eigenvalues from projected pseudo-gradients, with each diagonal entry:
\(\Lambda_{ii} = \mathbb{E}\left[ \left( (Q_A \otimes Q_Y) D_\theta \right)_i^2 \right]\), this captures the true variances of pseudo-gradients projected onto the Kronecker eigenbasis.

To compute the damped IHVP for influence functions:
\[
(G + \lambda I)^{-1} v \approx (Q_A \otimes Q_Y)(\Lambda + \lambda I)^{-1}(Q_A \otimes Q_Y)^\top v
\]

This correction preserves the efficiency of K-FAC while improving curvature approximation fidelity.


\section{Approximation Error Analysis}

Let the true inverse-Hessian-vector product (IHVP) denoted as:
\[
r^\star = (G + \lambda I)^{-1} v
\]
We compare the two approximations:
\begin{itemize}
    \item $r_{\text{LiSSA}}$ from LiSSA (Iterative, Stochastic)
    \item $r_{\text{EK-FAC}}$ from EK-FAC (Eigendecomposition-based)
\end{itemize}

\subsection{LiSSA Approximation Error Analysis}

LiSSA approximates the inverse Hessian-vector product $(G + \lambda I)^{-1} {v}$ recursively as:
\[
{r}_j = {v} + (I - \alpha(\tilde{G} + \lambda I)) {r}_{j-1}, {r}_0 = {v}
\]
    
For a matrix $A$, if $\|I - \alpha A\| < 1$, according to the Neumann series expansion, then:
\[
A^{-1} = \alpha \sum_{j=0}^{\infty} (I - \alpha A)^j
\]
Letting $A = G + \lambda I$, we have:
\[
(G + \lambda I)^{-1} = \alpha \sum_{j=0}^{\infty} (I - \alpha(G + \lambda I))^j
\]

LiSSA only uses the first $J+1$ terms, so the truncated sum is:
\[
r_{\text{LiSSA}}^{(J)} = \alpha \sum_{j=0}^J (I - \alpha(G + \lambda I))^j v
\]
while the ``ideal" solution is:
\[
r^\star = (G + \lambda I)^{-1} v = \alpha \sum_{j=0}^{\infty} (I - \alpha(G + \lambda I))^j v
\]

The error is the sum of the neglected terms:
\[
r^\star - r_{\text{LiSSA}}^{(J)} = \alpha \sum_{j=J+1}^{\infty} (I - \alpha(G + \lambda I))^j v
\]
This is a geometric series in operator form. Recall:
\[
\sum_{j=k}^{\infty} M^j = M^k (I - M)^{-1},\quad \text{for} \quad \|M\| < 1
\]
where $M = I - \alpha(G + \lambda I)$.

So,
\[
\sum_{j=J+1}^{\infty} M^j = M^{J+1}(I - M)^{-1}
\]
Therefore,
\[
r^\star - r_{\text{LiSSA}}^{(J)} = \alpha M^{J+1}(I - M)^{-1} v
\]

Applying the norm and sub-multiplicativity:
\[
\|r^\star - r_{\text{LiSSA}}^{(J)}\| \leq \alpha \|M^{J+1}\| \cdot \|(I - M)^{-1}\| \cdot \|v\|
\]

If $\|M\| < 1$, then
\[
\|(I - M)^{-1}\| \leq \frac{1}{1 - \|M\|}
\]
by the Neumann series for the matrix inverse. 
Plug in $M = I - \alpha(G + \lambda I)$:
\[
\|r^\star - r_{\text{LiSSA}}^{(J)}\| \leq 
\alpha \frac{ \| (I - \alpha(G + \lambda I))^{J+1} \| }
     { 1 - \| I - \alpha(G + \lambda I) \| } \cdot \|v\|
\]

LISSA convergence speed is slow and depends on the spectral radius $\rho(I - \alpha(G + \lambda I))$, which typically requires hundreds to thousands of Hessian-vector products (HVPs), and the stochasticity in estimating $G$ via mini-batches increases variance and error.

\subsection{EK-FAC Approximation Error Analysis}

EK-FAC approximates the Gauss-Newton Hessian $G$ using a Kronecker-factorized eigendecomposition and fitted eigenvalues:
\[
G \approx (Q_A \otimes Q_S) \Lambda (Q_A \otimes Q_S)^\top
\]

For the damped inverse Hessian-vector product, we have:
\[
(G + \lambda I)^{-1} {v} \approx (Q_A \otimes Q_S)
\left[
    \mathrm{diag} \left( \frac{1}{\lambda_i + \lambda} \right)
\right]
{(Q_A \otimes Q_S)^\top {v}}
\]

where $\lambda_i$ are the fitted eigenvalues obtained from the projected pseudo-gradients, $Q = Q_A \otimes Q_S$. 
Suppose the true matrix \( G \) and its EK-FAC approximation \( \hat{G} \) have the following eigendecomposition:
\[
G = Q \Lambda_{\text{true}} Q^\top
\]
\[
\hat{G} = Q \Lambda_{\text{EK}} Q^\top
\]
where \( Q \) is an orthonormal matrix of eigenvectors (\( Q^\top Q = I \)), and \( \Lambda_{\text{true}} \), \( \Lambda_{\text{EK}} \) are diagonal matrices of eigenvalues.

By the spectral theorem, for any vector \( v \):
\[
(G + \lambda I)^{-1} v = Q (\Lambda_{\text{true}} + \lambda I)^{-1} Q^\top v
\]
\[
(\hat{G} + \lambda I)^{-1} v = Q (\Lambda_{\text{EK}} + \lambda I)^{-1} Q^\top v
\]

Let \( r^\star = (G + \lambda I)^{-1} v \), \( r_{\text{EK-FAC}} = (\hat{G} + \lambda I)^{-1} v \).

Their difference is:
\[
r^\star - r_{\text{EK-FAC}} = \left[ Q (\Lambda_{\text{true}} + \lambda I)^{-1} Q^\top - Q (\Lambda_{\text{EK}} + \lambda I)^{-1} Q^\top \right] v
\]
\[
= Q \left[ (\Lambda_{\text{true}} + \lambda I)^{-1} - (\Lambda_{\text{EK}} + \lambda I)^{-1} \right] Q^\top v
\]

Let \( M = (\Lambda_{\text{true}} + \lambda I)^{-1} - (\Lambda_{\text{EK}} + \lambda I)^{-1} \), then
\[
r^\star - r_{\text{EK-FAC}} = Q M Q^\top v
\]

Taking norms and using the property \( \|A v\| \leq \|A\| \cdot \|v\| \) (operator norm):
\[
\|r^\star - r_{\text{EK-FAC}}\| = \| Q M Q^\top v \| \leq \| Q M Q^\top \| \cdot \|v\|
\]

Since \( Q \) is orthonormal, \( \|Q M Q^\top\| = \|M\| \):

So, the approximation error bound simplifies to:
\[
\|r^\star - r_{\text{EK-FAC}}\| \leq \| M \| \cdot \|v\|
\]
\[
= \left\| (\Lambda_{\text{true}} + \lambda I)^{-1} - (\Lambda_{\text{EK}} + \lambda I)^{-1} \right\| \cdot \|v\|
\]

This error only depends on how well the EK-FAC eigenvalues $\Lambda_{\text{EK}}$ approximate the true spectrum $\Lambda_{\text{true}}$, which is fixed once EK-FAC is fit. As there is no iterative refinement needed, the approximation error is significantly smaller than that of LiSSA for large models.

\section{Experiments}

\paragraph{Experiment Setup}  
We conduct a series of experiments across multiple configurations to evaluate the efficacy of influence functions for data attribution. Our primary objective is to assess whether influence-based methods can reliably identify training examples that significantly impact a model's prediction on a given test input. The experiments focus on image classification tasks, encompassing both low-complexity and high-complexity settings. Specifically, we evaluated on simple convolutional neural network (CNN) on MNIST~\cite{lecun2010mnist} and FashionMNIST~\cite{xiao2017fashion} datasets. For more complex evaluations, we consider ResNet50~\cite{he2016deep} with Flowers102~\cite{nilsback2008automated} and ViT-B/16~\cite{dosovitskiy2020image} on Food101~\cite{bossard14}.

\subsection{Influential Training Points Identification}
\begin{figure*}[ht]
  \centering
  \includegraphics[width=0.95\textwidth]{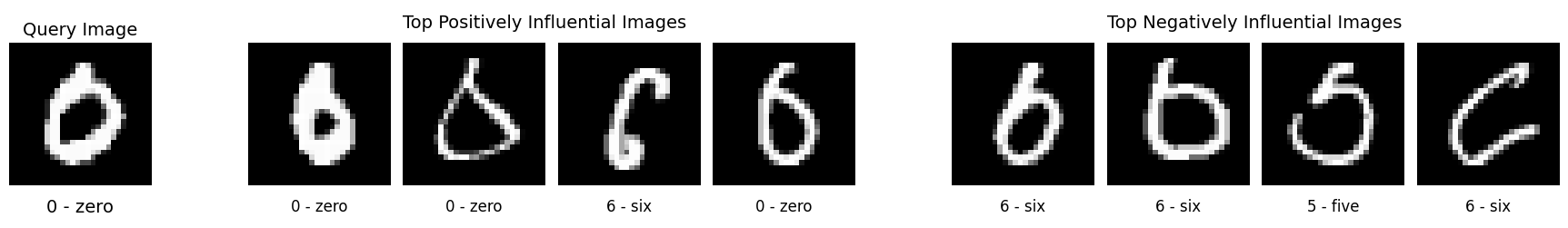}
  \includegraphics[width=0.95\textwidth]{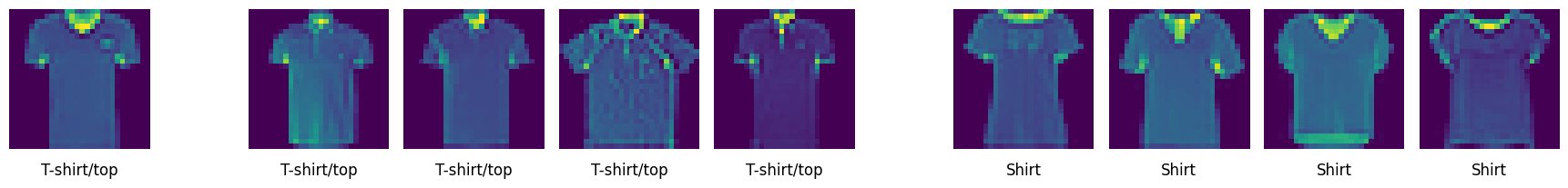}
  \includegraphics[width=0.95\textwidth]{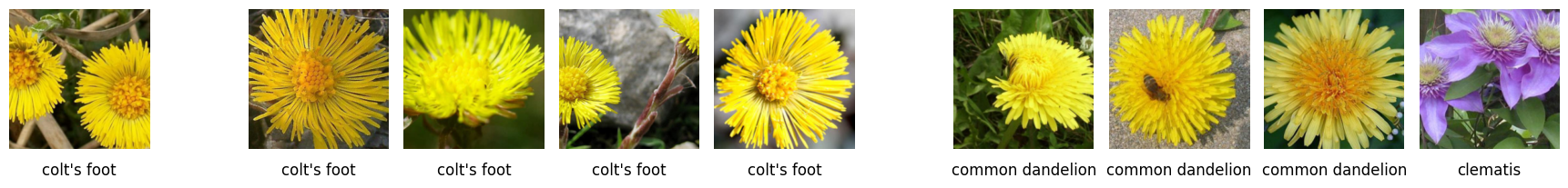}
  \includegraphics[width=0.95\textwidth]{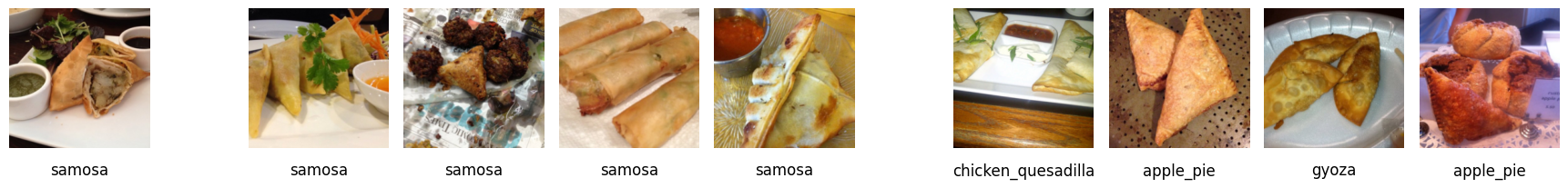}
  \caption{Most influential training samples identification regarding each query image.}
  \label{fig:1}
\end{figure*}

To examine the interpretability of influence functions, we analyze both positively and negatively influential training examples concerning selected test (query) instances. Qualitative results, presented in Fig.~\ref{fig:1}, provide visual evidence of the influence function's ability to identify relevant training samples across varying domains and model complexities.

\paragraph{MNIST and FashionMNIST (Simple CNN)}  
On the MNIST dataset, the influence function successfully identifies training examples with high morphological similarity to the test digit as positively influential. For instance, a query image of the digit ``0'' is most positively influenced by other well-formed ``0'' digits and occasionally by ``6''s due to visual similarity. Conversely, negatively influential examples include structurally dissimilar or ambiguous digits such as ``5'' or malformed ``0''s, which may confuse the classifier. This behavior illustrates the method's capability to capture both intra-class support and inter-class interference. Similarly, in the FashionMNIST setting—where the task involves distinguishing between categories of clothing—the influence function demonstrates sensitivity to subtle inter-class variations. For a test image labeled ``T-shirt/top,'' the top positively influential examples are visually consistent instances of T-shirts, while negatively influential examples include clothing items with differing silhouettes or necklines. These findings indicate that influence functions remain effective even under visually abstract and low-resolution scenarios. Additional qualitative examples are provided in Appendix Figs. 1 and 2.

\paragraph{Flowers102 (ResNet50)}  
On the more visually complex Flowers102 dataset, we observe a consistent pattern. A query image labeled ``colt’s foot'' is positively influenced by training samples that share similar color palettes and floral structures. Negatively influential samples include ``common dandelion'' and ``clematis,'' both of which share superficial visual traits but belong to different categories. This suggests that influence functions not only recover intra-class consistency but also expose sources of inter-class confusion, thereby offering fine-grained interpretability in high-resolution natural image classification. Additional results are presented in Appendix Fig.3.

\paragraph{Food101 (ViT-B/16)}  
In the Food101 experiments, influence scores again align with human-interpretable visual similarities. For a test image labeled ``samosa'', the top positively influential training examples belong to the same class, exhibiting common visual features such as triangular shape, golden-brown coloring, and plating presentation. Negatively influential examples include ``gyoza'' and ``apple pie,'' which, while visually similar, are categorically distinct. These results further highlight the influence function's utility in exposing class boundaries and model uncertainty in fine-grained recognition tasks. Additional qualitative examples are available in Appendix Fig.4.

\begin{figure*}[ht]
  \centering
  \includegraphics[width=0.95\textwidth]{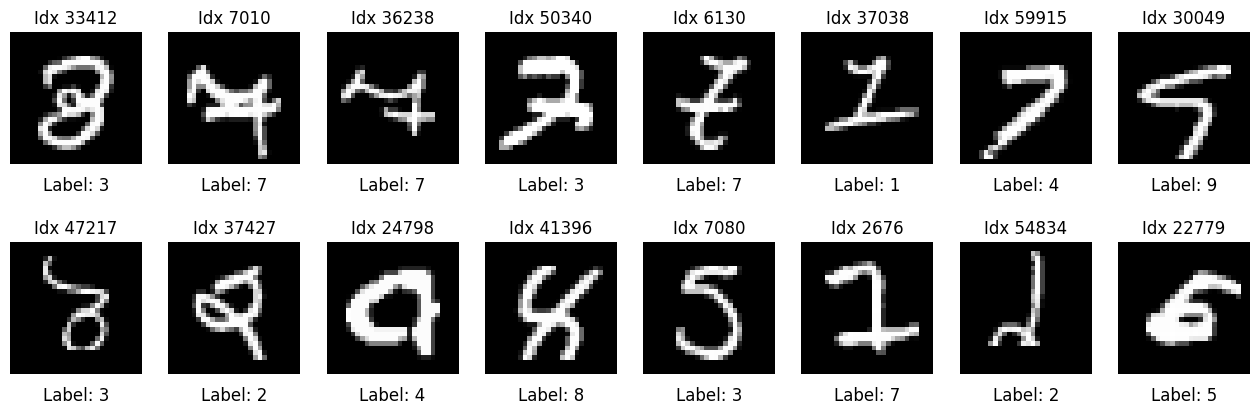}
  \caption{Mislabeled training samples retrieval from raw MNIST dataset.}
  \label{fig:2}
\end{figure*}

\subsection{Mislabeled Data Detection}

To evaluate the influence of the samples, we consider a scenario where humans need to inspect the training dataset quality, given that real-world data is bound to be noisy, and the bigger the dataset becomes, the more difficult it will be for humans to look for. The influence of a training point on itself is defined as Self-influence \cite{schioppa2022scaling}, which measures how up-weighting a training point $z$ contributes to reducing its loss during training. Namely:

\begin{equation}
\begin{aligned}
\text{Self-Influence}(z)
& =\mathcal{I}_{\text{loss}}(z,z) \\
& \approx -\nabla_{\theta} L\left(z, \theta\right)^{\top}
H_{\theta}^{-1} \nabla_{\theta} L(z, \theta)
\end{aligned} 
\end{equation}


Mislabeled training examples often exhibit strong self-influence, meaning they significantly reduce their own loss when upweighted, indicating the model is overfitting or bending excessively to accommodate them. This behavior, typical of outliers or mislabeled data, contrasts with clean examples, which usually have low self-influence as they are well-supported by the model and other data. By ranking training samples based on their self-influence scores in descending order, we can effectively identify mislabeled instances for human inspection. 

\begin{figure}[ht]
    \centering
    \includegraphics[width=0.9\linewidth]{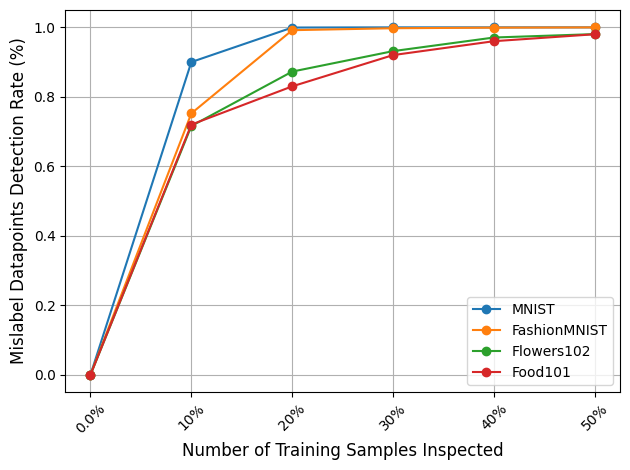}
    \caption{Mislabeled data detection performance with self-influence score.}
    \label{fig:mislabeled_detection}
\end{figure}

\paragraph{Quantitative Evaluation}
In this experiment, we evaluated the effectiveness of influence functions in detecting mislabeled samples of the training set. To simulate the real world mislabelling errors, we first construct a corrupted dataset with 10\% of the training data's labels randomly changed to another incorrect label and then retrain the model on the altered dataset. After retraining, self-influence scores were computed for each sample among the corrupted training set. By ranking samples according to these scores, we assessed detection accuracy at various inspection intervals (ranging from 10\% to 50\%) and measured the proportion of correctly identified corrupted samples among the top-ranked candidates. The experiment result can be seen from Fig. \ref{fig:mislabeled_detection}. Which demonstrates the potential of influence function for automatic identification of mislabeled data in large-scale dataset.

\paragraph{Qualitative Evaluation} 
Here we implement the self-influence calculation on the original uncorrupted MNIST dataset. Then we rank the samples according to the score and check the Top-20 samples manually, and found 16 of them are wrongly labeled or ambiguous. 
As can be seen from Fig.\ref{fig:2}, visual inspection reveals that many of these examples are ambiguous, highly distorted, or bear strong resemblance to digits of classes different from their annotated label. 
For instance:
\begin{itemize}
    \item Idx 33412 and 50340 (Label: 3): These resemble malformed or cursive “8”s more than canonical “3”s.
    \item Idx 7010 and 36238 (Label: 7): These exhibit nonstandard stroke patterns that look more like “4” or “9”.
    \item Idx 54834 (Label: 2): This sample is barely legible and appears more like a “1” or an outlier stroke artifact.
\end{itemize}

The majority of the retrieved samples show deviations from their labeled class and are plausible candidates for human mislabeling or inclusion error. Such cases are precisely the kinds of inconsistencies that can degrade model generalization, particularly in low-resource or noisy real-world scenarios. This experiment demonstrates that self-influence offers a practical, unsupervised criterion for prioritizing annotation audit in large-scale datasets. Compared to random inspection or simpler heuristics (e.g., loss magnitude), self-influence provides a theoretically grounded signal rooted in the model’s training dynamics. It captures how the model internalizes each training point relative to the rest of the dataset, making it particularly suitable for surfacing outliers and label noise.

\subsection{Influence Estimate Quality Evaluation}

\paragraph{Linear Datamodeling Score (LDS)}  
LDS \cite{park2023trak} evaluates the quality of data attribution methods by measuring how well they approximate the counterfactual effect of modifying subsets of training data on model outputs. It assumes linearity in attributions, i.e., the attribution for a subset is the sum of individual attributions. The process of calculating LDS is illustrated below:

\begin{enumerate}
    \item \textbf{Attribution-Based Prediction:}  
    Given a data attribution method \( \tau(z, Z) \), which scores training points in set \( Z \) with respect to a test point \( z \), the counterfactual prediction for a subset \( S \subset Z \) is defined as:
    \[
    g_\tau(z, S; Z) := \sum_{i: z_i \in S} \tau(z, Z)_i,
    \]

    \item \textbf{Per Sample LDS:}  
    LDS measures the Spearman rank correlation between true model outputs and attribution-based predictions across \( M \) random subsets \( \{S_j\}_{j=1}^M \):
    \[
    LDS(\tau, z) := \rho\left( \left\{f(z; \theta^\star(S_j)) \right\}_{j=1}^M, \left\{ g_\tau(z, S_j; Z) \right\}_{j=1}^M \right),
    \]
    where \( f(z; \theta^\star(S_j)) \) is the model output for \( z \) trained on subset \( S_j \), and \( \rho \) is Spearman correlation \cite{wissler1905spearman}. 

    \item \textbf{Subset-Averaged LDS:}  
    The overall LDS is the average across test subset examples:
    \[
    LDS(\tau) := \frac{1}{|\mathcal{Z}|} \sum_{z \in \mathcal{Z}} LDS(\tau, z)
    \]
\end{enumerate}

\begin{table}[ht]
\small
\centering
\begin{tabular}{lcccc}
\hline
\textbf{Task} & MNIST & FashionMNIST & Flowers102 & Food101 \\
\hline
LDS & 0.50 & 0.47 & 0.46 & 0.43 \\
\hline
\end{tabular}
\caption{LDS across different settings. Higher values indicate better alignment between attribution-based and true model behavior. We chose $M=100$, sub-sampling rate of $\alpha=0.5$ in our experiments, and the final LDS is averaged over 2000 test points randomly sampled from the testset.}
\end{table}

\section{Conclusion and Future Work}

In this work, we presented a comprehensive study on influence functions for data attribution in deep learning. Through theoretical analysis, algorithmic advancements, and empirical evaluation, we demonstrated the effectiveness in identifying impactful training samples and detecting mislabeled data across various-scale datasets and model architectures. Despite their conceptual elegance and utility, influence functions face challenges in scalability, robustness under non-convex optimization, and adaptation to modern training regimes. Looking forward, we identify two promising directions to extend influence-based techniques toward machine unlearning \cite{bourtoule2021machine}, aimed at efficiently mitigating the effects of harmful or mislabeled samples in trained models:

\paragraph{Sample Removing Based Unlearning} Once harmful or biased training samples (denoted as $D_{\text{forget}}$) are identified, their influence can be approximately removed from the model parameters $\theta$ using a single Newton update:

\[
    \begin{aligned}
        \theta_{\text{new}}
        & =\theta + \frac{1}{n}\sum_{z \in D_{\text{forget}}}H_{\theta}^{-1}\nabla_{\theta}L(z,\theta)
    \end{aligned}
\]

\paragraph{Label Repairing Based Unlearning} Instead of removing a mislabeled point, we can correct its label and apply a counterfactual correction to the model parameters:

\[
\theta_{\text{new}}=\theta + \frac{1}{n}\sum_{z \in D_{\text{forget}}}H_{\theta}^{-1}(\nabla_{\theta}L(\tilde{z},\theta)- \nabla_{\theta}L(z,\theta))
\]

Where $\tilde{z}$ demotes the label-repaired samples of $z$. This update nudges the model toward treating the corrected label as if it had been used during training. These proposed approaches avoid full retraining and allow efficient removal of unwanted influences.


\bibliography{aaai2026}


\appendix
\section{Relevant Derivation and Proof}

\subsection{The Derivation of $\mathcal{I}_{\text {params }}(z)$}
\label{proof:1}
Define the empirical risk:

\[
R(\theta) = \frac{1}{n} \sum_{i=1}^n L(z_i, \theta)
\]

Let \(\hat{\theta}\) minimize \(R(\theta)\), i.e.,

\[
\nabla R(\hat{\theta}) = 0
\]

Define the perturbed objective with upweighting a data point \(z\):

\[
\hat{\theta}_{\epsilon, z} = \arg\min_\theta \left\{ R(\theta) + \epsilon L(z, \theta) \right\}
\]

Since \(\hat{\theta}_{\epsilon,z}\) minimizes the perturbed loss, it satisfies:

\begin{equation} 
    0 = \nabla R(\hat{\theta}_{\epsilon,z}) + \epsilon \nabla L(z, \hat{\theta}_{\epsilon,z}) 
    \label{eq:A.1}
\end{equation}

Let \(\Delta_\epsilon = \hat{\theta}_{\epsilon,z} - \hat{\theta}_{0,z}\). Apply a first-order Taylor expansion:

\[
\nabla R(\hat{\theta}_{\epsilon,z}) \approx \nabla R(\hat{\theta}) + \nabla^2 R(\hat{\theta}) \Delta_\epsilon
\]

\[
\nabla L(z, \hat{\theta}_{\epsilon,z}) \approx \nabla L(z, \hat{\theta})+ \nabla^2 L(z,\hat{\theta}) \Delta_\epsilon
\]

Substitute into Eq. \eqref{eq:A.1}:

\[
0 \approx \left[ \nabla R(\hat{\theta}) + \epsilon \nabla L(z, \hat{\theta}) \right]
+ \left[ \nabla^2 R(\hat{\theta}) + \epsilon \nabla^2 L(z, \hat{\theta}) \right] \Delta_\epsilon
+ o(\|\Delta_\epsilon\|)
\]

We ignore the higher-order term \(o(\|\Delta_\epsilon\|)\) and $o(\epsilon)$, noting that \(\nabla R(\hat{\theta}) = 0\)

\[
0 \approx \epsilon \nabla L(z, \hat{\theta}) + \nabla^2 R(\hat{\theta}) \Delta_\epsilon
\]

We rearrange the above equation and solve for $\Delta_\epsilon$:

\[
\Delta_\epsilon \approx - \nabla^2 R(\hat{\theta})^{-1} \nabla L(z, \hat{\theta}) \epsilon
\]

Where \(H_{\hat{\theta}} = \nabla^2 R(\hat{\theta})\), $d \hat{\theta}_{\epsilon,z}= \frac{\hat{\theta}_{\epsilon,z} - \hat{\theta}_{0,z}}{\epsilon} = d \Delta_\epsilon$ hence:

\[
\mathcal{I}_{\text{params}}(z) \stackrel{\text { def }}{=} \left. \frac{d \hat{\theta}_{\epsilon,z}}{d\epsilon} \right|_{\epsilon=0} = \left. \frac{d \Delta_\epsilon}{d\epsilon} \right|_{\epsilon=0} = - H_{\hat{\theta}}^{-1} \nabla L(z, \hat{\theta})
\]

\subsection{The Proof of $\hat{\theta}_{-z}-\hat{\theta} \approx-\frac{1}{n} \mathcal{I}_{\text {params }}(z)$}
\label{proof:2}

\begin{proof}{(\(\hat{\theta}_{-z}-\hat{\theta} \approx-\frac{1}{n} \mathcal{I}_{\text {params }}(z)\))}
\begin{align*}
    \hat{\theta}_{-z} 
    & \stackrel{\text { def }}{=} \arg \min_{\theta \in \Theta} \frac{1}{n}\sum_{z_{i} \neq z} L\left(z_{i}, \theta\right) \\
    & = \arg\min_{\theta \in \Theta}\frac{1}{n}(\sum_{i=1}^{n} L\left(z_{i}, \theta\right) - L(z,\theta)) \\
    \hat{\theta}_{-\frac{1}{n}, z} 
    &\stackrel{\text { def }}{=} \arg \min _{\theta \in \Theta} \frac{1}{n} \sum_{i=1}^{n} L\left(z_{i}, \theta\right)-\frac{1}{n} L(z, \theta) \\
    &= \arg\min_{\theta \in \Theta}\frac{1}{n}(\sum_{i=1}^{n} L\left(z_{i}, \theta\right) - L(z,\theta)) \\
    &\hat{\theta}_{-z} = \hat{\theta}_{-\frac{1}{n}, z} \\
    & \hat{\theta} \stackrel{\text { def }}{=} \arg \min _{\theta \in \Theta} \frac{1}{n} \sum_{i=1}^{n} L\left(z_{i}, \theta\right)\\
    \hat{\theta}_{0, z} 
    &\stackrel{\text { def }}{=} \arg \min _{\theta \in \Theta} \frac{1}{n} \sum_{i=1}^{n} L\left(z_{i}, \theta\right)-0*L(z, \theta) \\
    &= \arg \min _{\theta \in \Theta} \frac{1}{n} \sum_{i=1}^{n} L\left(z_{i}, \theta\right)\\
    & \hat{\theta} = \hat{\theta}_{0, z} 
\end{align*}

\[
\begin{aligned}
    \hat{\theta}_{-z}-\hat{\theta} 
    & = \hat{\theta}_{-\frac{1}{n}, z} - \hat{\theta}_{0, z} \\
    & = -\frac{1}{n}*{\frac{\hat{\theta}_{-\frac{1}{n}, z} - \hat{\theta}_{0, z}}{(-\frac{1}{n})-0}} \\
    & = -\frac{1}{n}*{\left.\frac{d \hat{\theta}_{\epsilon, z}}{d \epsilon}\right|_{\epsilon=0}}
\end{aligned}
\]

\end{proof}

\begin{figure*}[ht]
  \centering
  \includegraphics[width=0.95\textwidth]{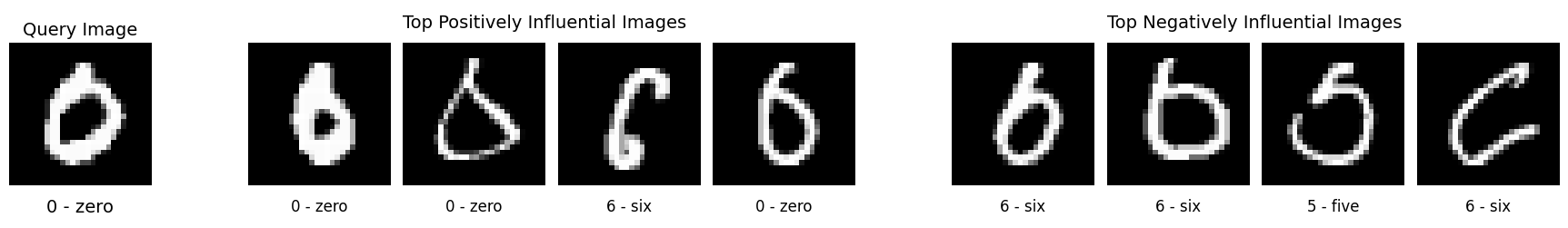}
  \includegraphics[width=0.95\textwidth]{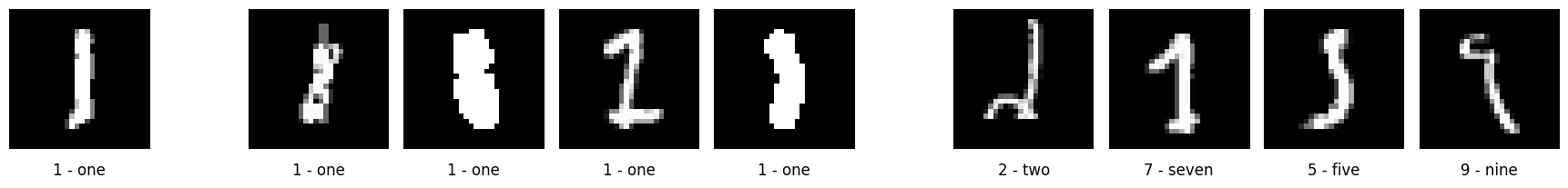}
  \includegraphics[width=0.95\textwidth]{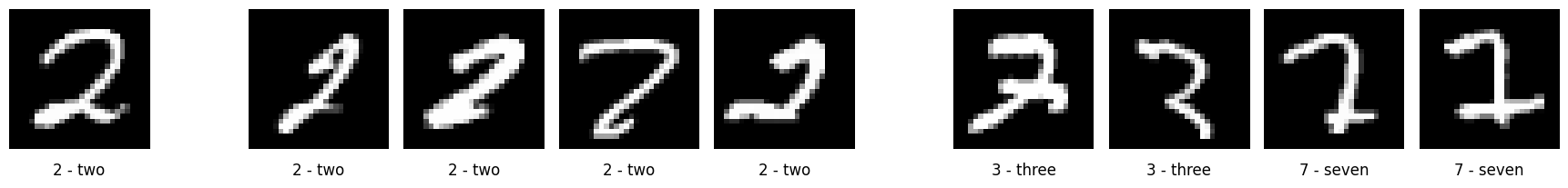}
  \includegraphics[width=0.95\textwidth]{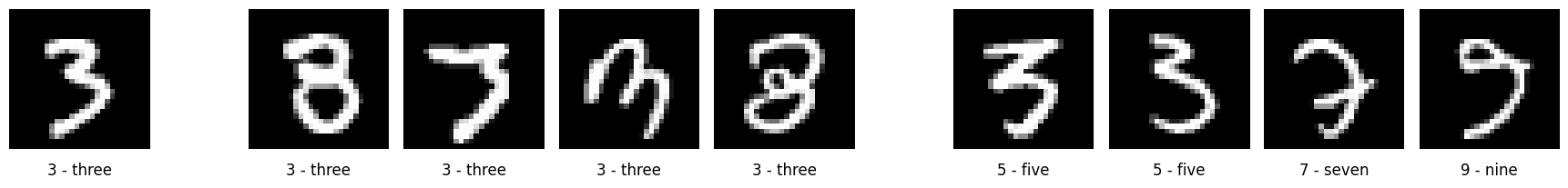}
  \includegraphics[width=0.95\textwidth]{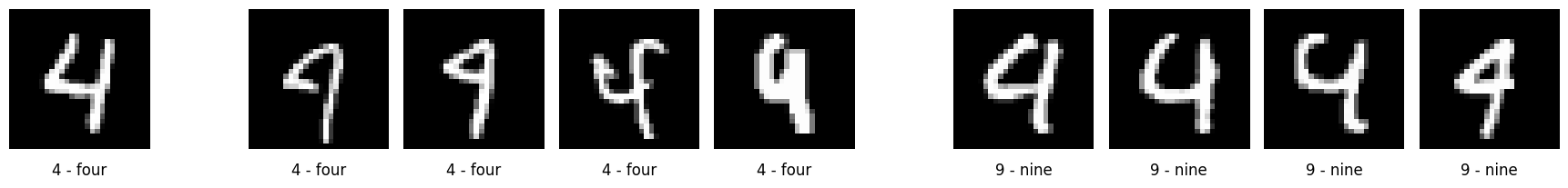}
  \includegraphics[width=0.95\textwidth]{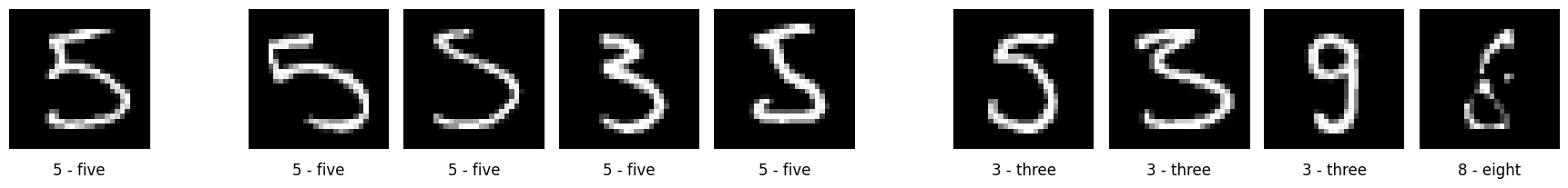}
  \includegraphics[width=0.95\textwidth]{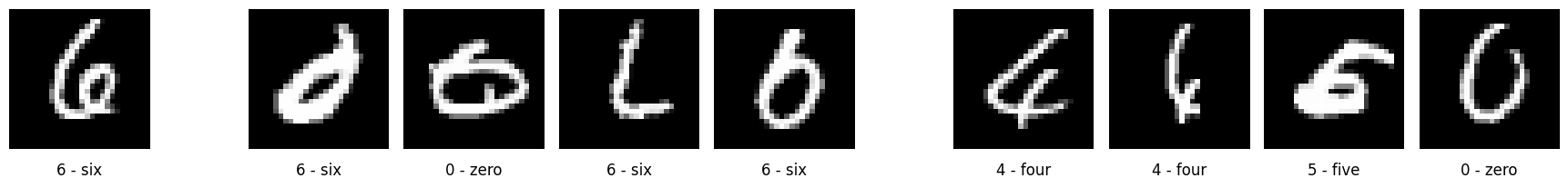}
  \includegraphics[width=0.95\textwidth]{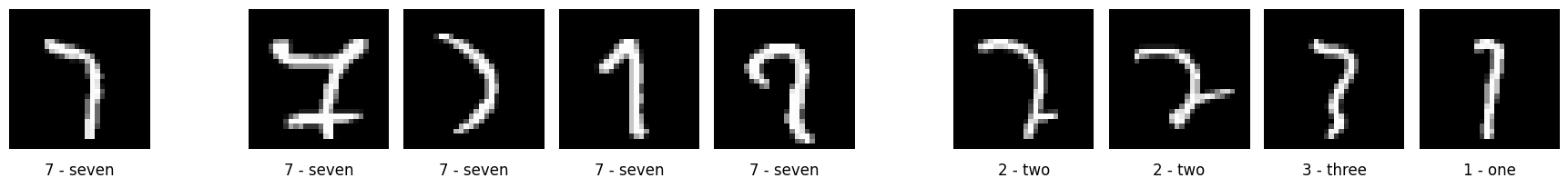}
  \includegraphics[width=0.95\textwidth]{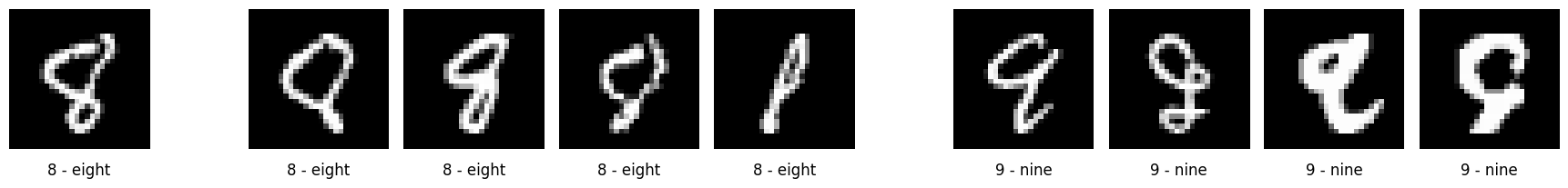}
  \includegraphics[width=0.95\textwidth]{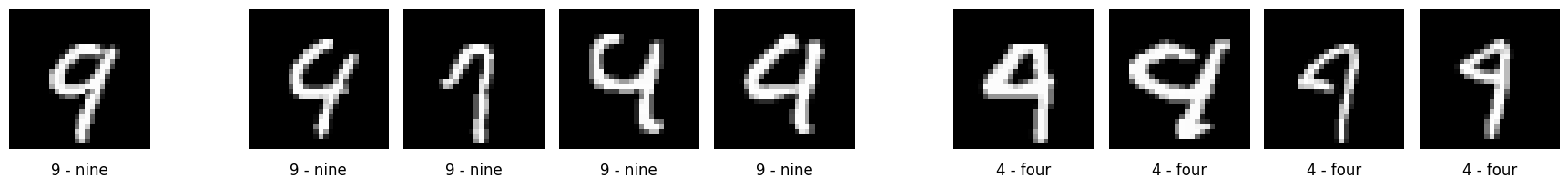}
  \caption{MNIST Influential Data Identification}
  \label{fig:ap_mnist}
\end{figure*}

\begin{figure*}[ht]
  \centering
  \includegraphics[width=0.95\textwidth]{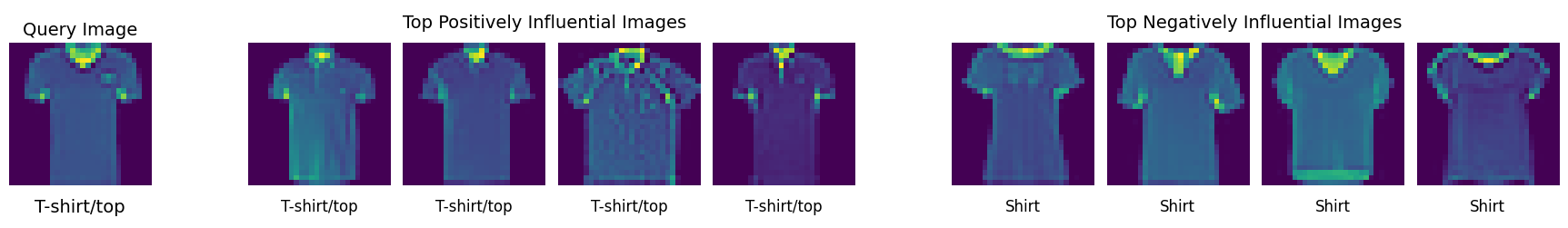}
  \includegraphics[width=0.95\textwidth]{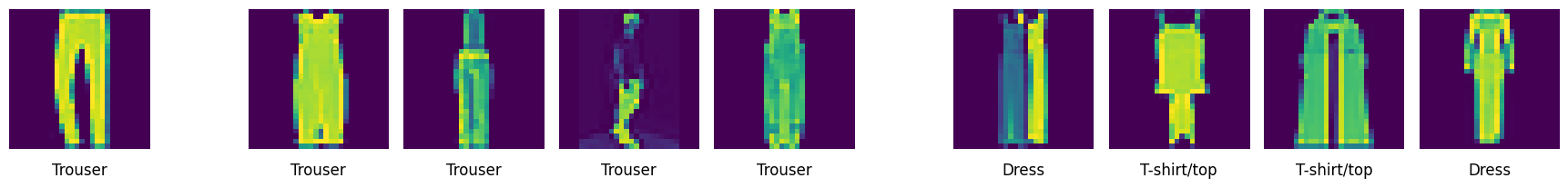}
  \includegraphics[width=0.95\textwidth]{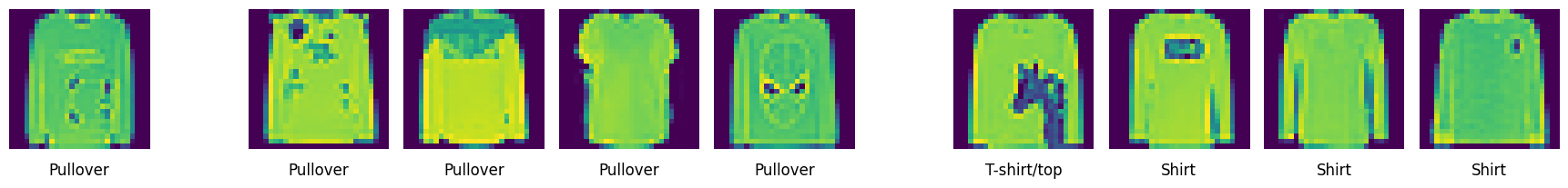}
  \includegraphics[width=0.95\textwidth]{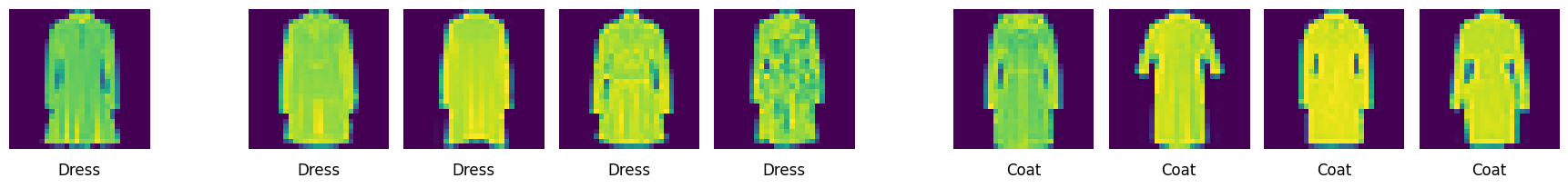}
  \includegraphics[width=0.95\textwidth]{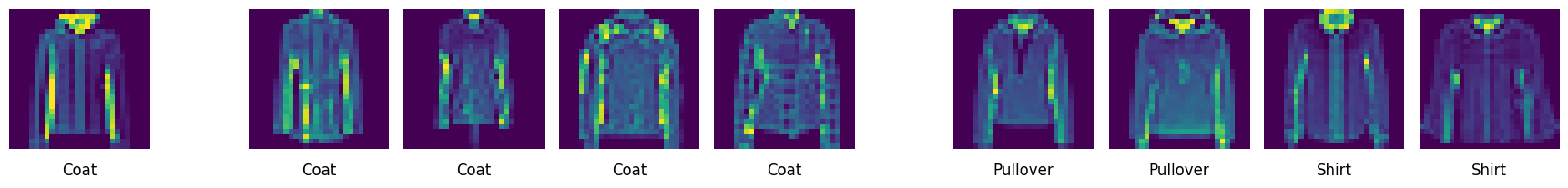}
  \includegraphics[width=0.95\textwidth]{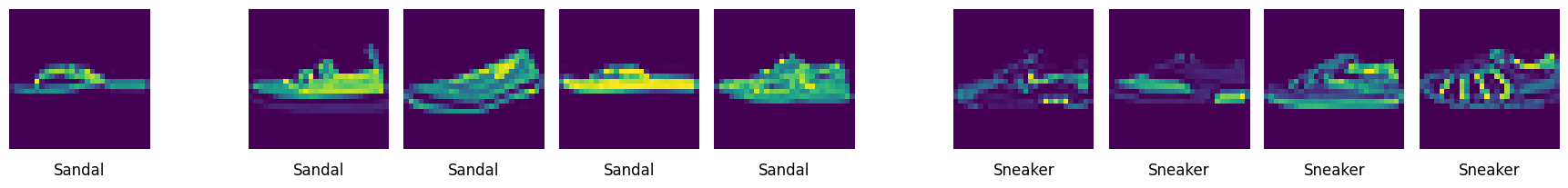}
  \includegraphics[width=0.95\textwidth]{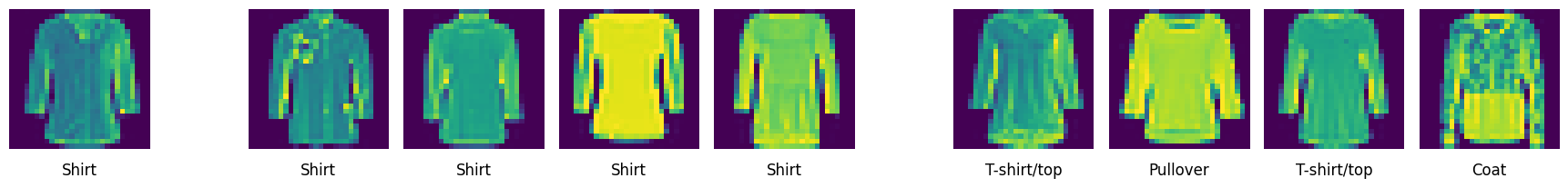}
  \includegraphics[width=0.95\textwidth]{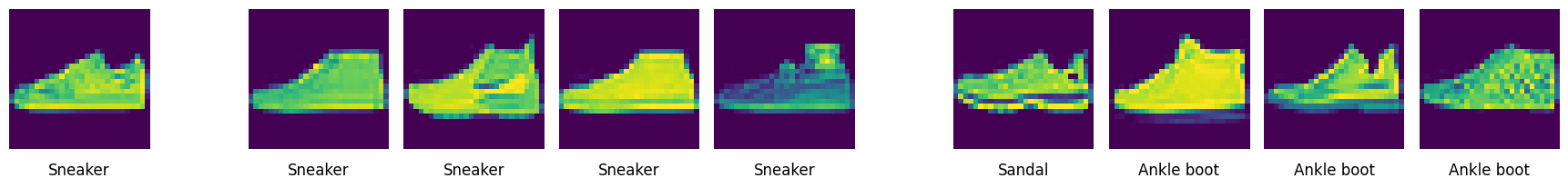}
  \includegraphics[width=0.95\textwidth]{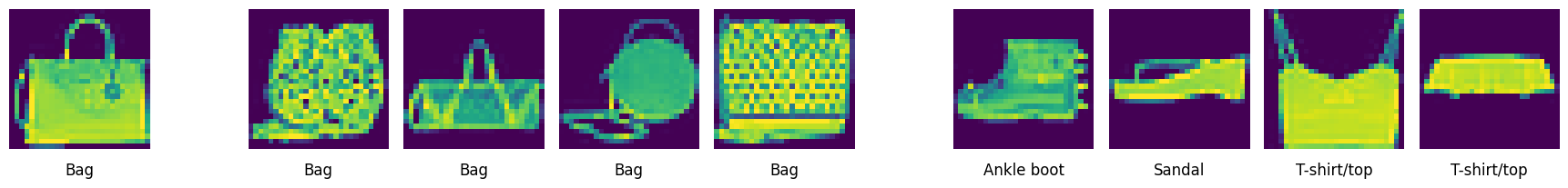}
  \includegraphics[width=0.95\textwidth]{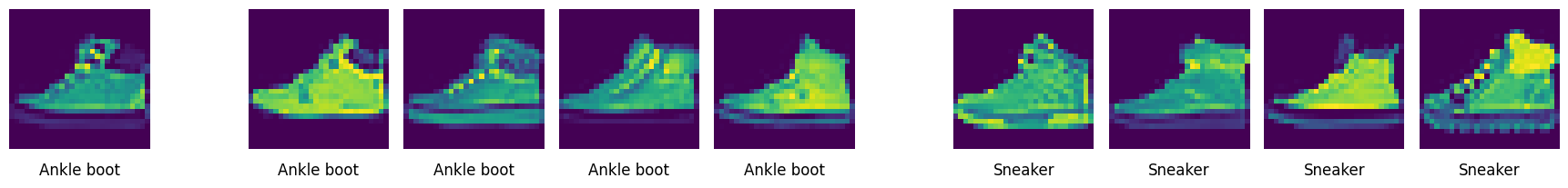}
  \caption{FashionMNIST Influential Data Identification}
  \label{fig:ap_fashion_mnist}
\end{figure*}

\begin{figure*}[ht]
  \centering
  \includegraphics[width=0.95\textwidth]{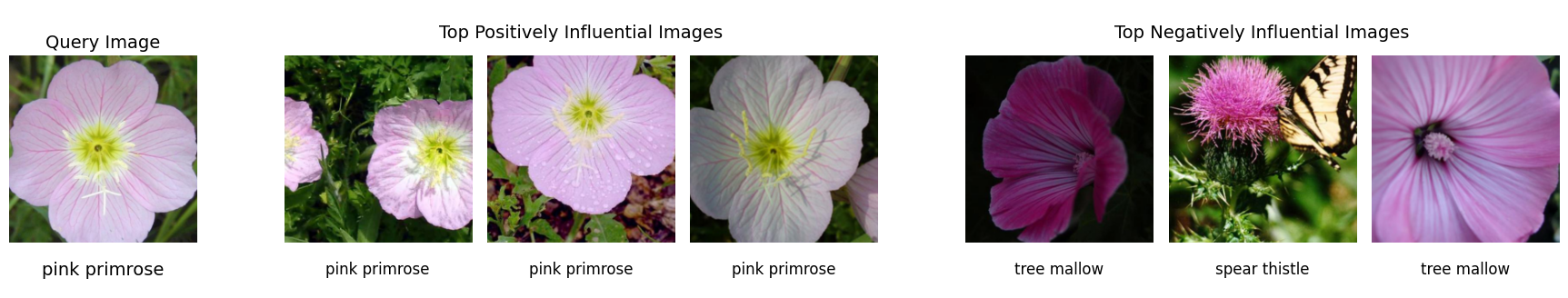}
  \includegraphics[width=0.95\textwidth]{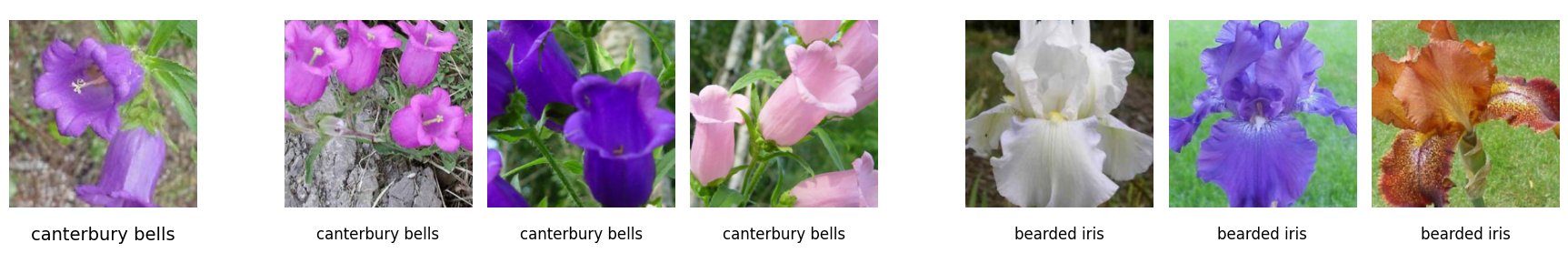}
  \includegraphics[width=0.95\textwidth]{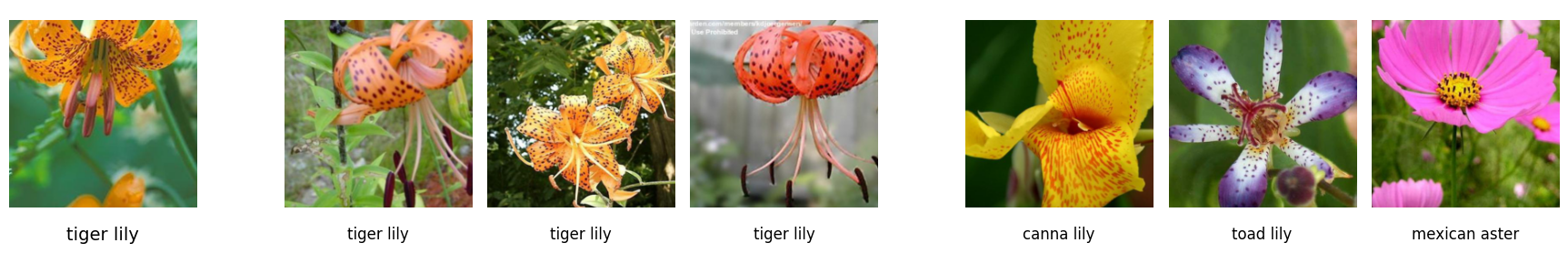}
  \includegraphics[width=0.95\textwidth]{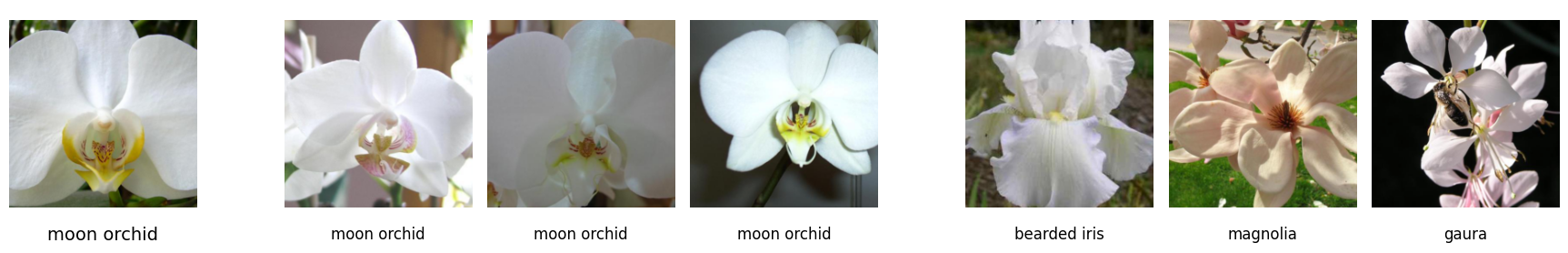}
  \includegraphics[width=0.95\textwidth]{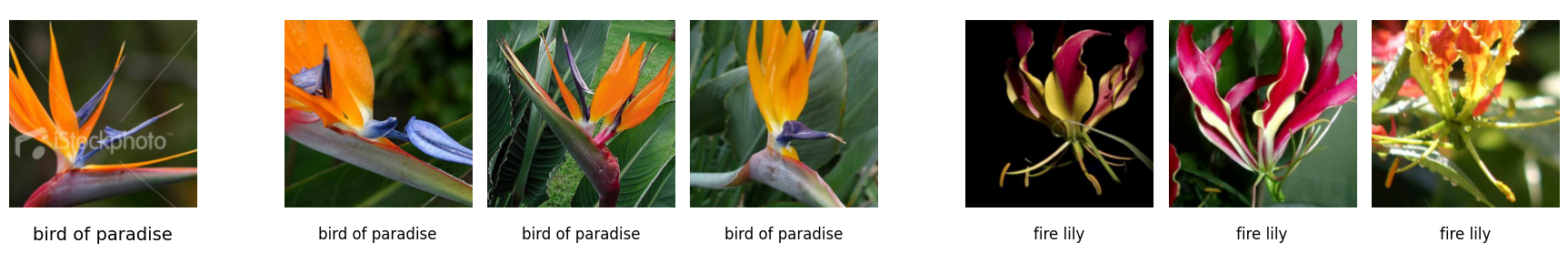}
  \includegraphics[width=0.95\textwidth]{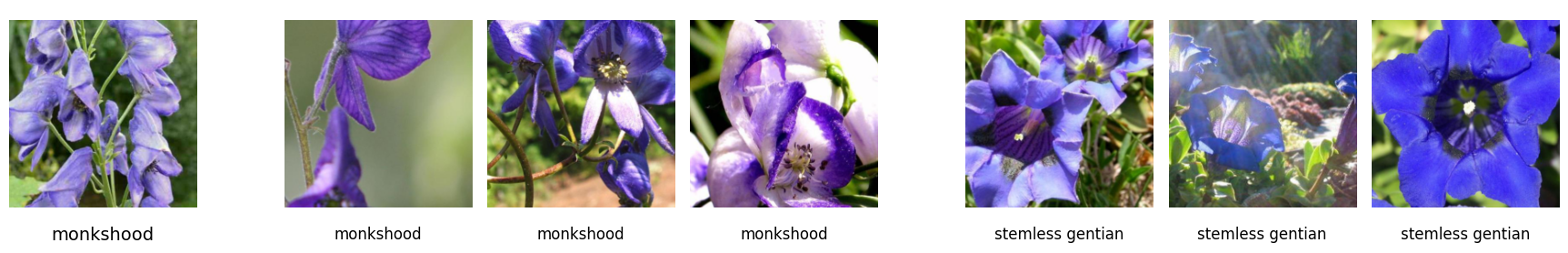}
  \includegraphics[width=0.95\textwidth]{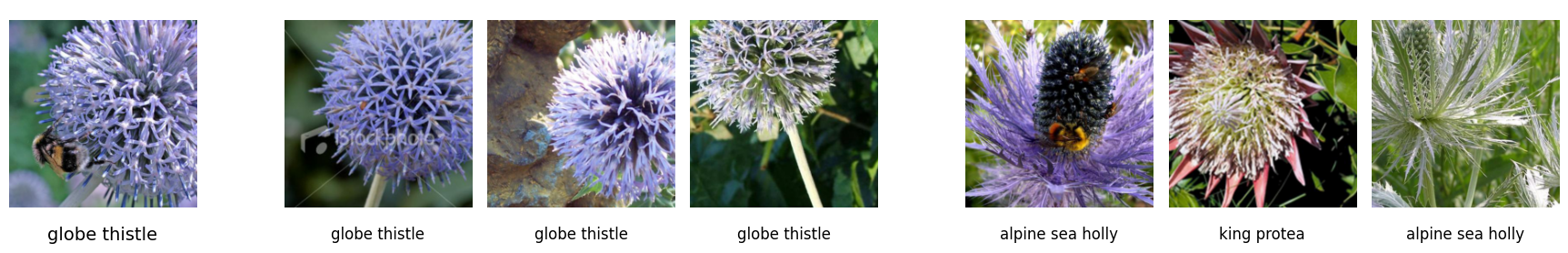}
  \includegraphics[width=0.95\textwidth]{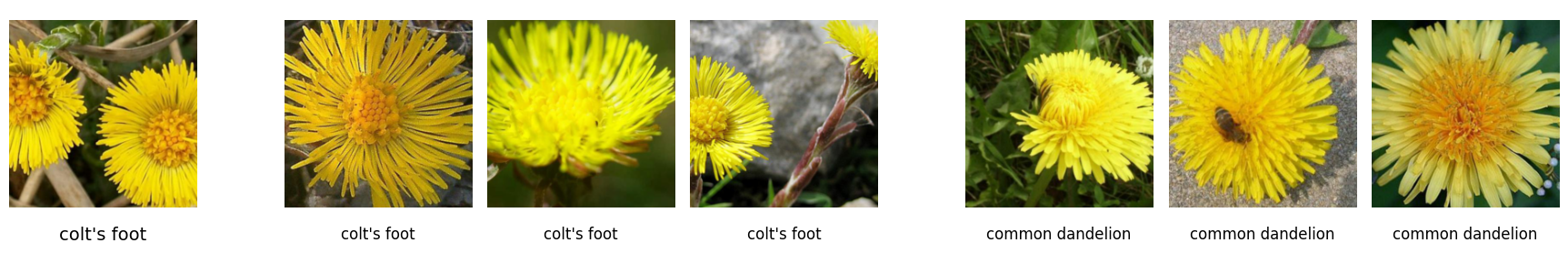}
  \caption{Flowers102 (ResNet50) Influential Data Identification}
  \label{fig:ap_flowers102}
\end{figure*}

\begin{figure*}[ht]
  \centering
  \includegraphics[width=0.95\textwidth]{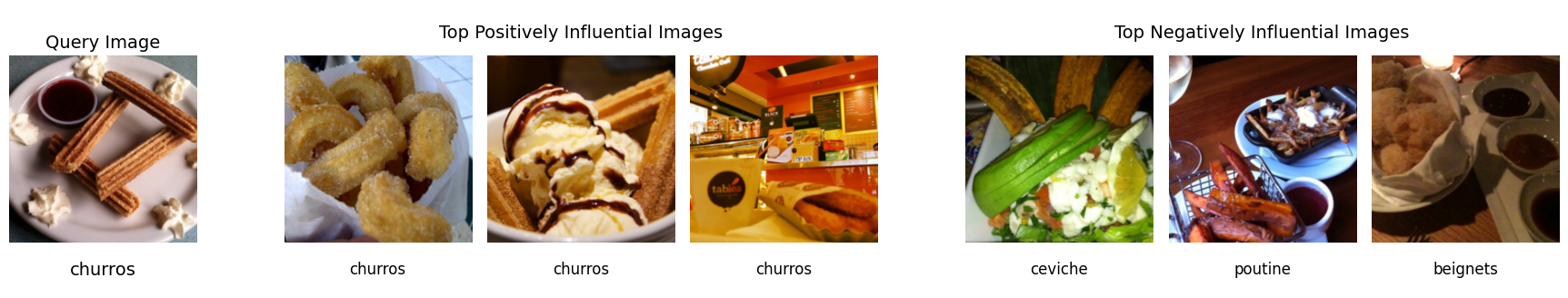}
  \includegraphics[width=0.95\textwidth]{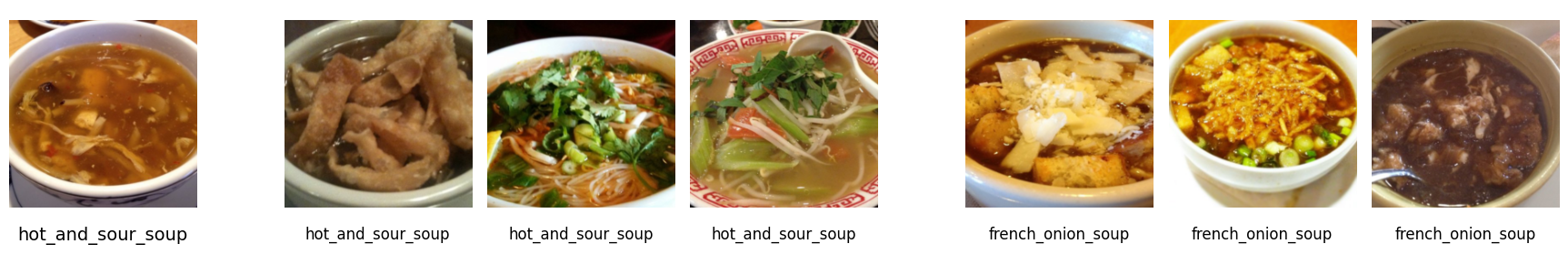}
  \includegraphics[width=0.95\textwidth]{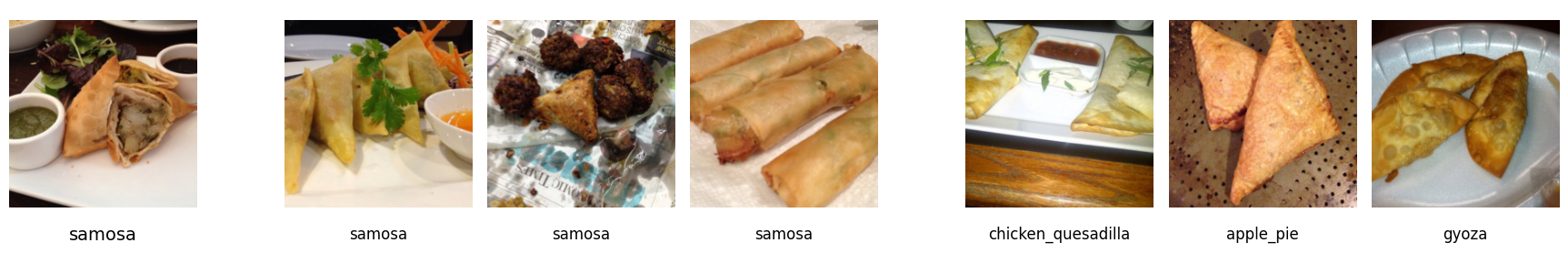}
  \includegraphics[width=0.95\textwidth]{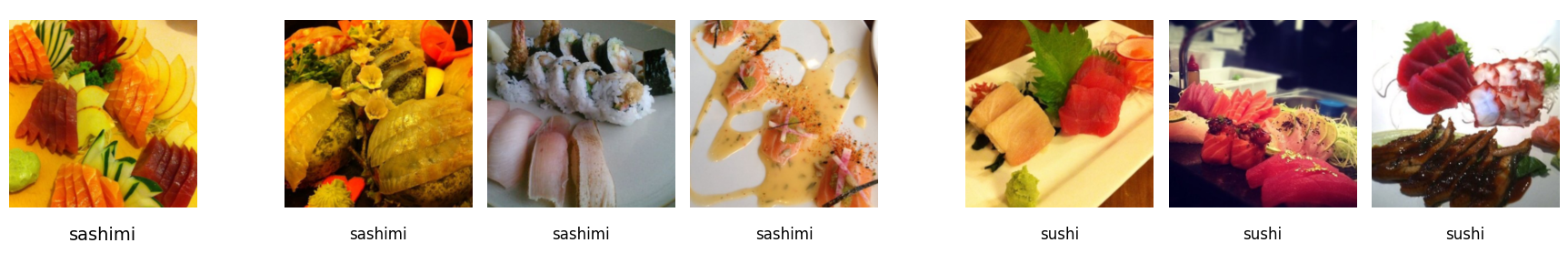}
  \includegraphics[width=0.95\textwidth]{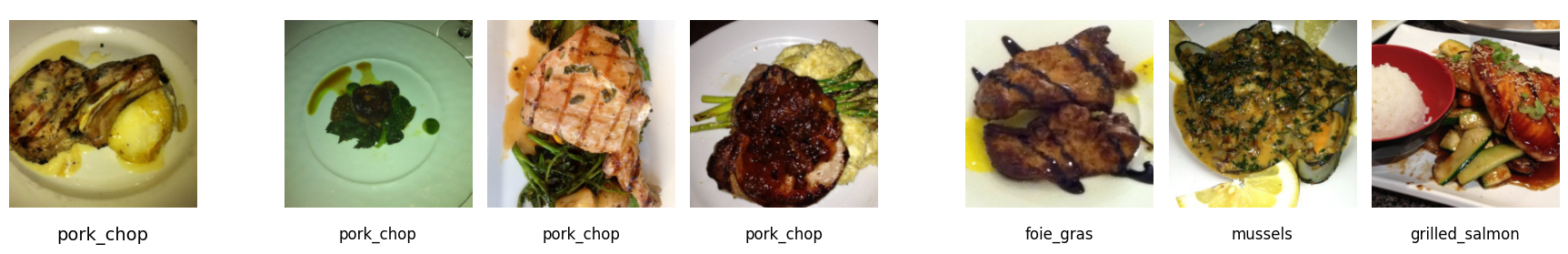}
  \includegraphics[width=0.95\textwidth]{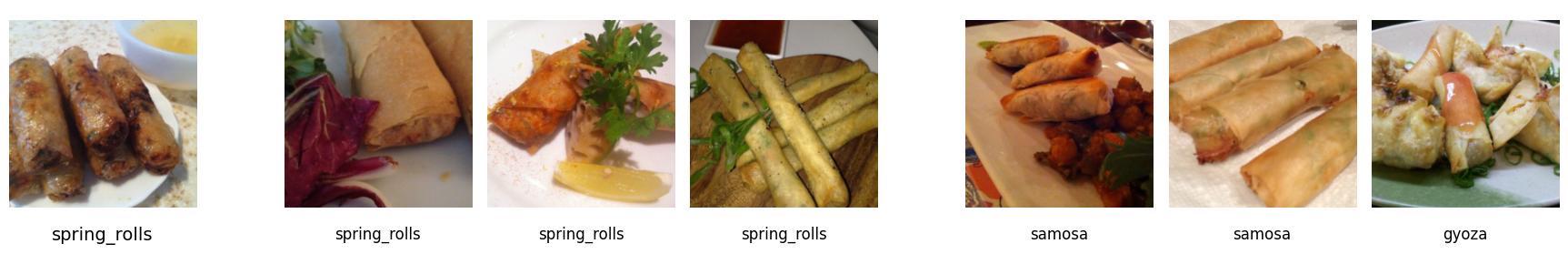}
  \includegraphics[width=0.95\textwidth]{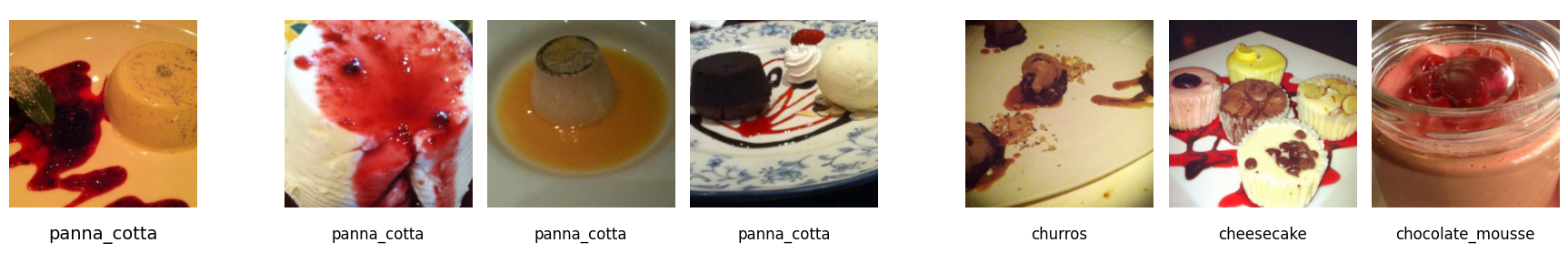}
  \includegraphics[width=0.95\textwidth]{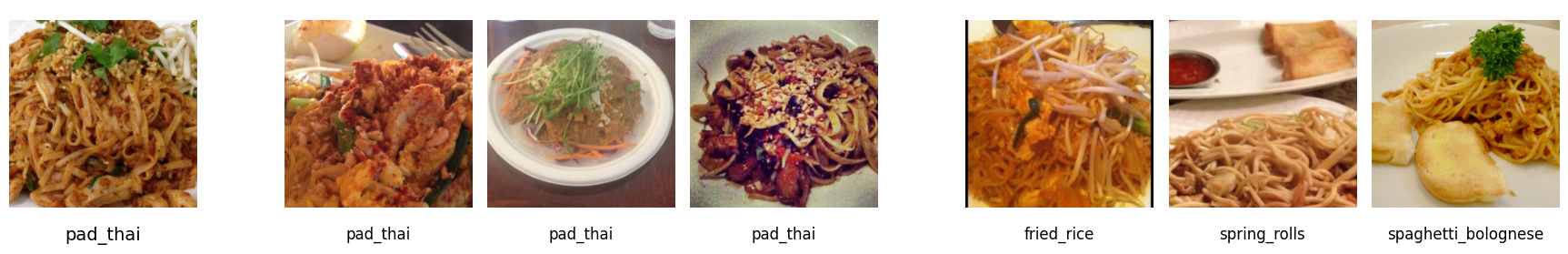}
  \caption{Food101 (ViT-B/16) Influential Data Identification}
  \label{fig:ap_food101}
\end{figure*}

\end{document}